# Transfer Learning for Wildlife Classification: Evaluating YOLOv8 against DenseNet, ResNet, and VGGNet on a Custom Dataset

Subek Sharma*, Sisir Dhakal*, and Mansi Bhavsar[@]

*Department of Electronics and Computer Engineering, Institute of Engineering (IOE) Pashchimanchal Campus, Tribhuvan University, Pokhara, Gandaki, Nepal, Corresponding author: subeksharma11@gmail.com

[@]Department of Computer Information Science, Minnesota State University, Mankato

*joint first authors.

**Abstract-** *This study evaluates the performance of various deep learning models, specifically DenseNet, ResNet, VGGNet, and YOLOv8, for wildlife species classification on a custom dataset. The dataset comprises 575 images of 23 endangered species sourced from reputable online repositories. We utilize transfer learning to fine-tune pre-trained models on the dataset, focusing on reducing training time and enhancing classification accuracy. The results demonstrate that YOLOv8 outperforms other models, achieving a training accuracy of 97.39'% and a validation F1-score of 96.50'%. These findings suggest that YOLOv8, with its advanced architecture and efficient feature extraction capabilities, holds great promise for automating wildlife monitoring and conservation efforts.*

***Keywords-** Convolutional Neural Network (CNN), Endangered Species Detection, Image Classification, Transfer Learning, YOLO.*

## 1. Introduction

Each living creature in the world plays a vital role in maintaining the balance of the ecosystem. However, activities like poaching and illegal trade, reduction of prey base, habitat loss and degradation, and human-wildlife conflict have led to a rapid decline in the number of some species, including critically endangered ones [1]. Currently, wildlife monitoring is done using camera traps that capture the images of animals, which are then analyzed and studied by humans. This is time-consuming, tedious, and prone to errors.

The introduction of various deep learning techniques and their use in computer vision shows optimistic results. Convolutional Neural Networks (CNNs), introduced by LeCun et al. in 1998 [2], have been a significant milestone in image classification tasks [3]–[5]. In these networks, images are fed through convolutional and pooling layers for feature extraction, followed by fully connected layers. Transfer learning, highlighted by Pan and Yang [6], is fine-tuning pre-trained models on large datasets for specified tasks that significantly reduce training time with increased performance.

Densely connected convolutional networks, popularized by Huang et al. in 2017 [7], and residual networks, put forward by He et al. in 2016 [8], are deep and complex networks that facilitate effective feature extraction. However, VGGNet, proposed by Simonyan and Zisserman in 2014 [9], is somewhat computationally expensive, although its results are excellent in some cases.YOLOv8 is an evolution in Redmon et al.'s object detection and has shown remarkable results in various computer vision tasks [10].

El Abbadi et al.[11] achieved a classification accuracy of 97.5% using a deep convolutional neural network model for the automated classification of vertebrate animals, thereby demonstrating the effectiveness of deep learning in animal recognition tasks. Similarly, Villa et al.[12] demonstrated in their study that very deep convolutional neural networks achieved 88.9% accuracy in a balanced dataset and 35.4% in an unbalanced one for automatic species identification in camera-trap images, marking a notable advancement in non-intrusive wildlife monitoring. In 2020 Ibraheam et al.[13] reported in their paper that their deep learning-based system achieved 99.8% accuracy in distinguishing between animals and humans, and 97.6% in identifying specific animal species, significantly improving safety in wildlife-human and wildlife-vehicle encounters.

Brust et al.[14] have illustrated that ResNet50 outperformed VGG16 and Inception v3 on a wildlife image dataset, with the highest accuracy of 90.3 %. It provided insights into the strengths and weaknesses of different CNN architectures for wildlife classification [15]. Similarly, Beery et al.[16] 2018 validated the same on a challenging dataset of wildlife images with occlusions, varying lighting conditions, and motion blur. They found the Faster R-CNN model achieved the highest accuracy of 88.7 %, indicating the importance of model robustness in real-world applications. Similarly, Yilmaz et al.[17] in 2021 demonstrated that the YOLOv4 algorithm achieved a high classification accuracy of 92.85 % for cattle breed detection, emphasizing its effectiveness in wildlife classification tasks. The research in [18] evaluates deep learning-based models, including SSD, YOLOv4, and YOLOv5, on the CUB-200-2011 dataset. The YOLOv4 model outperforms state-of-the-art methods with high accuracy and precision. Hung Nguyen et al.[19] achieved 96.6 % accuracy in detecting animal images and 90.4 % accuracy in species identification using deep learning, highlighting its potential for automatically monitoring wildlife.

This paper answers some of the questions regarding the selection of deep learning models for practical wildlife conservation tasks, such as which models provide the best accuracy and efficiency, how YOLOv8 compares to DenseNet, ResNet, and VGGNet, and what challenges and limitations exist in applying these models to wildlife conservation. The results show that YOLOv8 is best suited for automated wildlife monitoring with much better accuracy and efficiency than models such as DenseNet, ResNet, and VGGNet. This work will supply essential guidance to researchers and practitioners on the choice of appropriate models for endangered species conservation. Addressing these research questions is crucial as the current methodologies for wildlife monitoring are labor-intensive and error-prone. This work aims to fill this gap by systematically evaluating different deep-learning models and providing essential guidance to researchers and practitioners on the choice of appropriate models for endangered species conservation.

In this paper, we begin by reviewing related works and current methodologies in wildlife conservation. In Section 2, we describe the dataset and the preprocessing steps involved, as well as provide a detailed overview of the methodologies employed in this study. Following this, Section 3 elaborates on the performance and evaluation metrics used for the machine learning models. In Section 4, we present and discuss the experimental results. Finally, we conclude in Section 5 with a summary of the findings and suggestions for future work.

## 2. Methodology

### 2.1. *Dataset Description*

**2.1.1.** *Species Coverage:* The dataset includes 23 each carefully selected based on its conservation status and the need for monitoring. These species cover many endangered animals, including mammals, reptiles, and amphibians as shown in Table 1.

| Scientific Name | Common Name | Quantity |
| --- | --- | --- |
| Ailurus fulgens | Red Panda | 25 |
| Aonyx cinerea | Asian Small-Clawed Otter | 25 |
| Arctictis binturong | Binturong | 25 |
| Bubalus arnee | Wild Water Buffalo | 25 |
| Catopuma temminckii | Asian Golden Cat | 25 |
| Cervus duvaucelii | Swamp Deer (Barasingha) | 25 |
| Elephas maximus | Asian Elephant | 25 |
| Rhinoceros unicornis | Indian Rhinoceros | 25 |
| Indotestudo elongata | Elongated tortoise | 25 |
| Lutrogale perspicillata | Smooth-Coated Otter | 25 |
| Macaca assamensis | Assam Macaque | 25 |
| Manis pentadactyla | Chinese Pangolin | 25 |
| Melanochelys tricarinata | Three-keeled land turtle | 25 |
| Melursus ursinus | Sloth Bear | 25 |
| Neofelis nebulosa | Clouded Leopard | 25 |
| Panthera pardus | Leopard | 25 |
| Panthera tigris | Bengal Tiger | 25 |
| Prionailurus viverrinus | Fishing Cat | 25 |
| Ratufa bicolor | Black Giant Squirrel | 25 |
| Uncia uncia | Snow Leopard | 25 |
| Ursus thibetanus | Asian Black Bear | 25 |
| Varanus flavescens | Yellow Monitor Lizard | 25 |
| Viverra zibetha | Large Indian Civet | 25 |

Table 1: Different classes of species in the dataset with their common and scientific names.

2.1.2. ***Data Collection:*** The study began by collecting the data from different sources on the internet. Each class is represented by 25 filtered images, resulting in 575 images in the dataset. We maintained a balanced dataset to reduce the bias towards any particular species and to perform the balanced model training. The internet houses a huge amount of data, but finding and gathering useful ones is still challenging. To collect images of various animal species, we utilized online repositories like iNaturalist and ZooChat [20], [21]. The good thing about using such sites is their authenticity and fair use policy. For the same reason, we did not use the images shown in regular Google searches or try to automate the process.

## 2.2. *Data Preprocessing*

We divided the preprocessing of the image data into the following steps.

**2.2.1.** *Aspect Ratio Standardization:* We preprocessed all the images utilized for our study to have an aspect ratio of 1:1 and a resolution of 400 x 400. We added padding whenever required, ensuring we did not lose any information from the images during resizing.

**2.2.2.** *Data Normalization:* We normalized every image to increase accuracy and speed up the model's convergence. It also reduced the variance in the training data.

**2.2.3.** *Splitting:* We split the dataset into train (80%) and val (20%) sets. We used the split-folders Python package to split the dataset while maintaining the original distribution [22]. Babu et al.'s study examines the impact of image data splitting on the performance of machine learning models [23].

**2.2.4.** *Data Augmentation:* For better generalization, we increased the diversity of data by applying different augmentation techniques, as suggested by Shorten and Khoshgoftaar in 2019 [24]. Data augmentation improves the performance of deep learning systems by artificially inflating the size of their training dataset through various data manipulation methods, including rotation, flipping, translation, and other color-oriented transformations. Such transformations help the model become robust against overfitting to specific patterns in the training data and instead enable the model to generalize better on new data. Through augmentation, the model learns more informative features that are stable across various real-world variations. Hence, it becomes insensitive to the variations in input conditions. Table 2 displays the final set of parameters for data augmentation.

| Augmentation | Description |
| --- | --- |
| Shearing | Random shearing of the image by a maximum of 20%. |
| Zooming | Random zooming in or out of the image by a maximum of 20%. |
| Horizontal flip | Random horizontal flipping of the image. |
| Rotation | Randomly rotating the image by up to 20 degrees. |
| Width shifting | Randomly shifting the image horizontally by 10% of the image width. |
| Height shifting | Randomly shifting the image vertically by a maximum of 10% of the image height. |

**Table 2: Different augmentation parameters used for training.**

### 2.3. Convolutional Neural Network

Convolutional Neural Networks are a special kind of neural network designed to work with grid data like images. They learn by extracting the features from input data via convolution and pooling operations followed by fully connected layers [2]. CNNs have played a pivotal role in the advancement of computer vision and related tasks. They are especially effective in performing tasks such as image recognition, object detection, and classification [25]-[27].

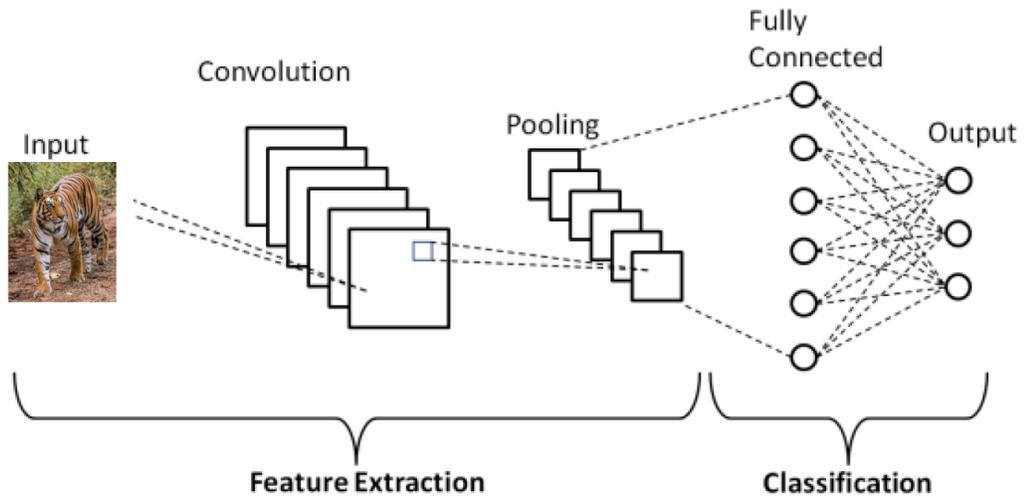

Figure 1: Convolutional Neural Network. [28]

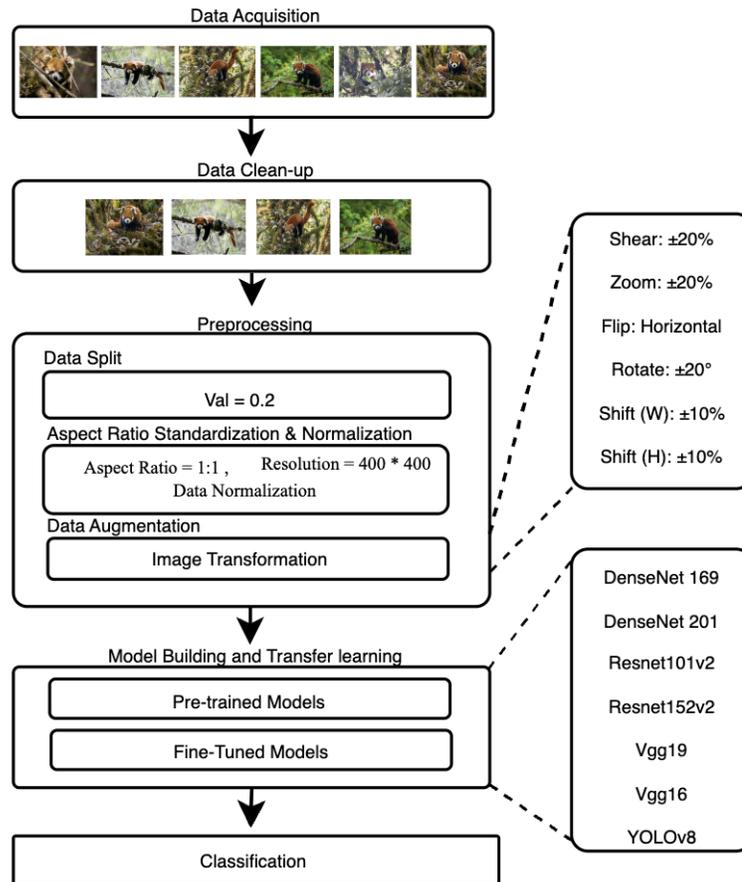

Figure 2: Methodology.

## 2.4. Transfer Learning

Transfer learning is an approach to deep learning that enables researchers and developers to use the previously trained model in a huge dataset and implement it in downstream tasks [3]. It is especially useful when we have limited data to train the model. Also, it significantly reduces the training time because the feature extraction part remains unchanged. However, it may fail when there is a significant mismatch between the target domain task and the source task.

These are some of the models we used to compare transfer learning with our dataset.

- **2.4.1.** ***DenseNet:*** It was introduced by Gao Huang and colleagues in 2017 [4]. It connects each layer with all other layers densely, meaning each layer receives input from the preceding layers. By efficiently cutting the parameters' requirements and boosting the network's ability by reusing the features extracted, it allows for a deeper network. However, the dense connectivity may lead to increased computational cost and memory consumption.

- **2.4.2.** ***ResNetV2:*** It was introduced by Kaiming He et al. in 2016 [5]. It utilizes residual blocks, which are shortcut connections that bypass one or multiple layers. It also allows the network to learn residual functions instead of direct mapping. This architecture supports very deep networks without degradation problems.

- **2.4.3.** ***VGG:*** It was introduced by Karen Simonyan and Andrew Zisserman in 2014 [6]. It is known for its simplicity and depth, achieved by stacking small 3*3 convolutional layers, increasing the model's depth and parameters. The max pooling layers follow the convolutional layers to handle the volume size and end with the fully connected layers.

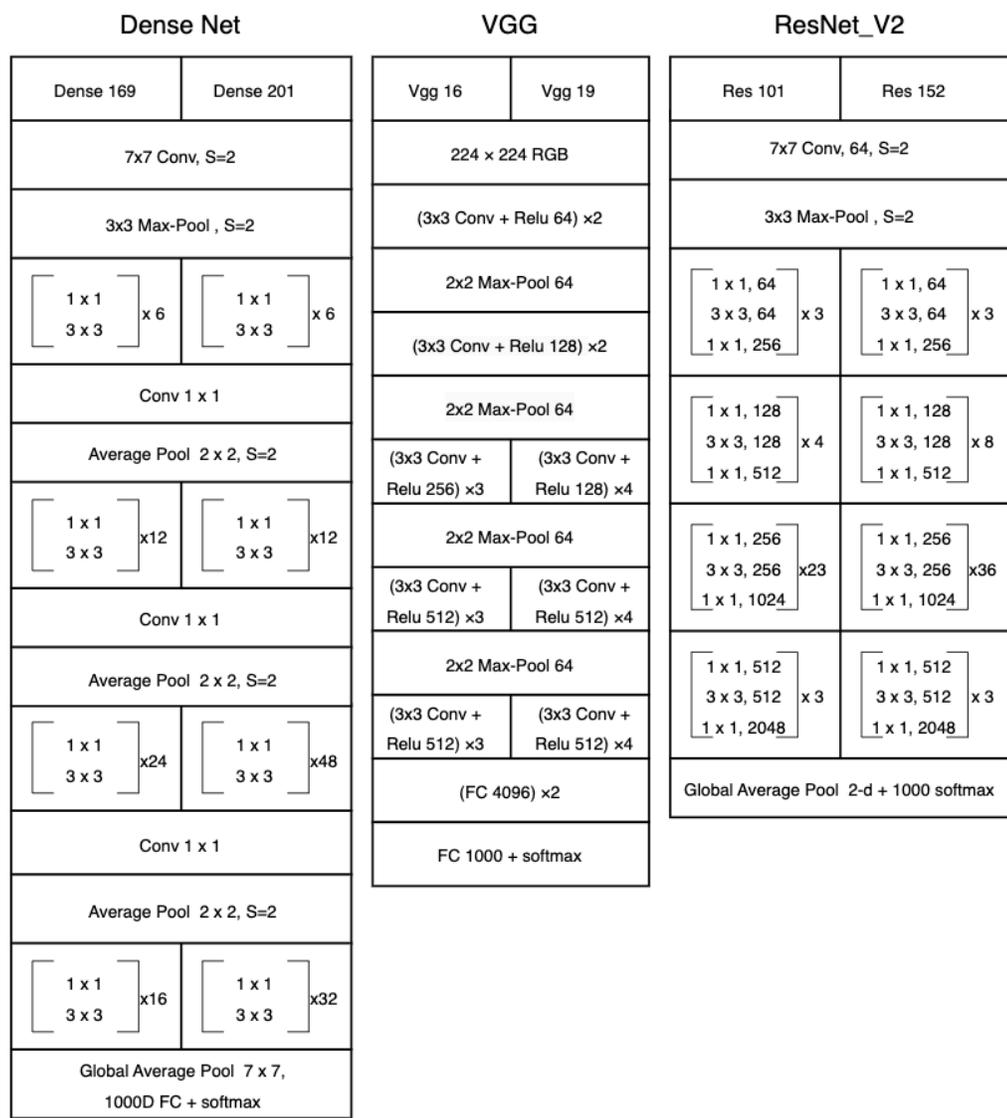

**Figure 3: Architecture of different models used in this study.**

- **2.4.4. YOLOv8:** YOLOv8 is built upon the object detection framework introduced by Joseph Redmon, with contributions from many researchers over successive versions. It enhances speed and accuracy through an advanced backbone architecture, refined loss functions, and anchor-free detections [28]. Figure 4 shows the architecture of the YOLOv8.

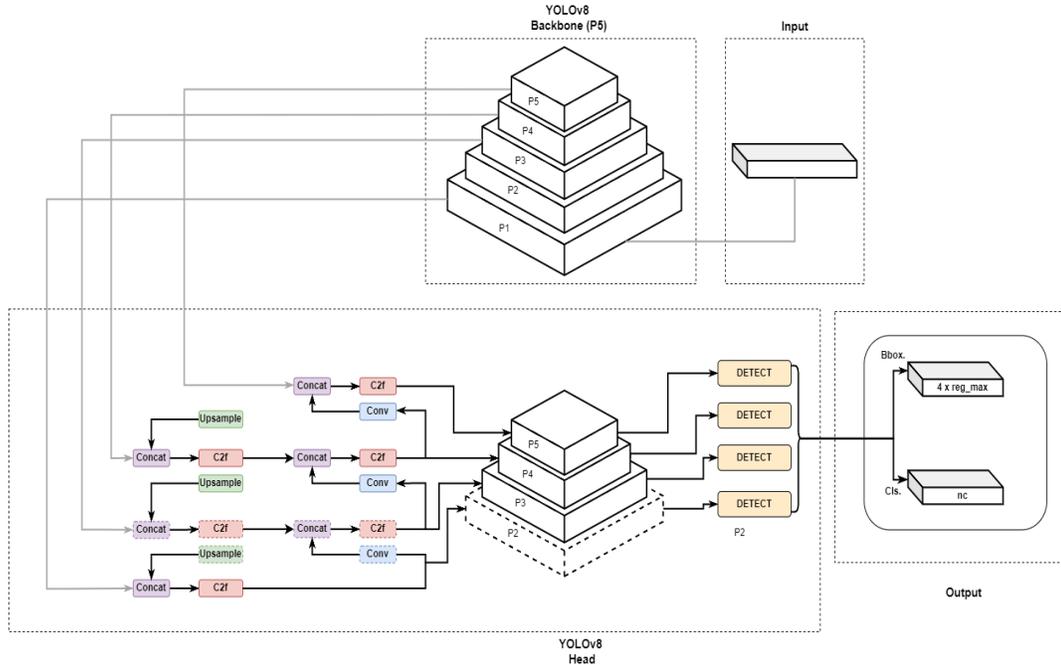

**Figure 4: Reference image for YOLOv8 architecture.**

## 2.5. Model building

We loaded the pre-trained models with their respective weights, made these weights untrainable(frozen), and replaced the last layer with custom, fully connected layers corresponding to the number of classes in our dataset. We added a GlobalAveragePooling2D layer to reduce the feature maps' spatial dimensions and prevent overfitting. This layer is followed by a Dense layer with 128 neurons and activated by ReLU to introduce non-linearity and learn more complex features. Finally, we added a Dense layer with 23 units activated by softmax to match the number of classes in our dataset, enabling the model to output the class probabilities, as shown in Figure 5. We then trained the models using Adam as the optimizer and cross-entropy as the loss function for 100 epochs with a batch size of 32 images. Meanwhile, we used a validation set to monitor the progress to avoid plateauing and adjusting the learning rate dynamically.

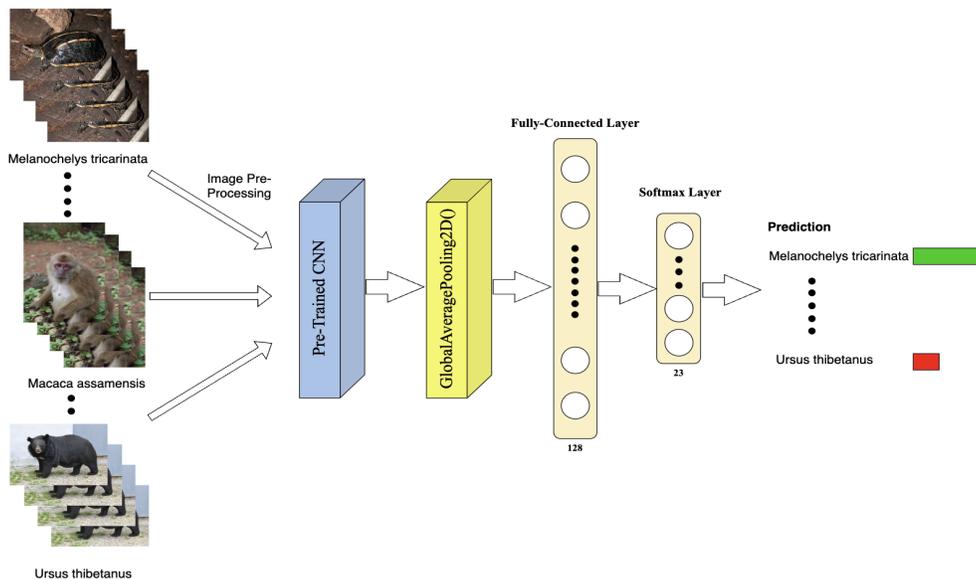

**Figure 5: Illustration of transfer learning.**

## 2.6. Hyperparameters

We experimented with multiple sets of hyperparameters, including learning rates, optimizers, batch sizes, and schedulers, thereby selecting the most favorable settings of hyperparameters that showed optimal performance. Table 3 lists the final set of hyperparameters.

### 2.6.1. Categorical Cross Entropy:
Categorical Cross Entropy (CCE) is used for multi-class classification tasks. It measures the difference between the true class labels and the predicted class probabilities. The formula sums the negative log-likelihood of the true class's predicted probability across all classes. We have used CCE as our loss function.

$$Categorical\ Entropy\ =\ -\sum_{neuron=1}^{classes} y_{true_{neuron}} * ln\left(y_{pred_{neuron}}\right) \tag{1}$$

### 2.6.2. Binary Cross Entropy for One Neuron:
Binary Cross Entropy (BCE) is used for binary classification tasks. The binary cross-entropy loss measures the dissimilarity between predicted probability distributions and the ground truth labels. The formula is the negative log-likelihood of the predicted probability if the true label is 1 and 1 minus the predicted probability if the true label is 0. The BCE is utilized in each output layer neuron and used in the YOLOv8 classification task.

$$BCE_{1_{neuron}} = -\left[y_{true} * ln\left(y_{predicted}\right) + \left(1 - y_{true}\right) * ln\left(1 - y_{predicted}\right)\right] \tag{2}$$

### 2.6.3. AdamW Optimizer:
AdamW stands for Adaptive Moment Estimation with Weight Decay. It is an extension of the Adam optimizer, including weight decay, to improve generalization by preventing overfitting. AdamW adjusts the learning rate based on the gradients' first and second moments and consists of a weight decay term to penalize large weights.

$$AdamW\ =\ \theta_{t+1} = \theta_t - \frac{\alpha}{\widehat{v}_t + \epsilon} * \widehat{m}_t - \alpha * weight\_decay * \theta_t \tag{3}$$

Where:

- $\theta_t$ is the parameter at time step t

- $\alpha$ is a learning rate

- $\widehat{m}_t$ is the biased first-moment estimate

- $\widehat{v}_t$ is the biased second raw moment estimate

- $\epsilon$ is a small constant to prevent division by zero

- weight_decay is the weight decay term.

| Hyperparameters | Value |
|---|---|
| Learning Rate | Adaptive |
| Batch Size | 32 |
| Epochs | 100 |
| Early Stopping | {ReduceLROnPlateau, Validation Accuracy} |
| Optimizers | {Adam, AdamW} |

**Table 3: Hyperparameters.**

## 3. Performance and Evaluation Metrics

The performance of the models is not examined by accuracy alone. In addition to accuracy, other metrics like f1-score, precision, recall, and loss were employed for the evaluation. Precision is an indicator of the accuracy of model predictions i.e., the ratio of the true positive predictions to the total number of positive predictions made by the model. The recall is an indicator of the model's ability to identify all the relevant classes i.e., ratio of true positive predictions to the total number of actual positive instances. The F1 score provides the single value for the evaluation of the model and is the harmonic mean of precision and recall. The loss gives insights into how well the model's prediction matches the true outcomes and is the difference between the predicted values and the actual values.

$$Accuracy = \frac{True\ Positives + True\ Negative}{Total\ number\ of\ sample\ data} \tag{4}$$

$$Precision = \frac{Total\ Number\ of\ True\ Positives}{True\ Positives + False\ Positives} \tag{5}$$

$$Recall = \frac{Total\ Number\ of\ True\ Positives}{True\ Positives + False\ Negatives} \tag{6}$$

$$F1\ score = 2\frac{Precision * Recall}{Precision + Recall} \tag{7}$$

## 4. Results And Discussion

### 4.1. Results

This study evaluated multiple deep-learning models for identifying endangered animal species from wildlife images. The results show the varying performance across the models. YOLOv8 outperformed the other models, and the DenseNet and Resnet models also did well, as their results were close to those of YOLOv8. In contrast, the VGG and Vanilla CNN models faced significant challenges. Table 4 shows the performances of different models.

| S.N | Model Name | Accuracy(%) | Loss | F1-Score(%) | Precision(%) | Recall(%) |
|---|---|---|---|---|---|---|
| 1 | DenseNet 169 | Train = 98.27<br>Val = 93.91 | Train = 0.0265<br>Val = 0.0482 | Train = 95.22<br>Val = 93.95 | Train = 99.46<br>Val = 98.94 | Train = 83.76<br>Val = 80.87 |
| 2 | DenseNet 201 | Train = 98.80<br>Val = 92.17 | Train = 0.0161<br>Val = 0.0386 | Train = 96.36<br>Val = 92.22 | Train = 99.95<br>Val = 94.62 | Train = 89.97<br>Val = 76.52 |
| 3 | Resnet 101 V2 | Train = 98.74<br>Val = 92.17 | Train = 0.0083<br>Val = 0.9217 | Train = 97.36<br>Val = 92.09 | Train = 99.42<br>Val = 98.00 | Train = 95.72<br>Val = 85.22 |
| 4 | Resnet 152 V2 | Train = 98.20<br>Val = 93.04 | Train = 0.0104<br>Val = 0.0271 | Train = 95.79<br>Val = 93.22 | Train = 99.19<br>Val = 97.14 | Train = 94.01<br>Val = 88.70 |
| 5 | VGG 16 | Train = 46.00<br>Val = 33.91 | Train = 0.5619<br>Val = 0.5994 | Train = 42.89<br>Val = 28.65 | Train = 0.00<br>Val = 0.00 | Train = 0.00<br>Val = 0.00 |
| 6 | VGG 19 | Train = 32.67<br>Val = 33.04 | Train = 0.6222<br>Val = 0.6449 | Train = 29.82<br>Val = 31.19 | Train = 0.00<br>Val = 0.00 | Train = 0.00<br>Val = 0.00 |
| 7 | YOLOV8 | Train = 97.39<br>Val = 99.13 | Train = 0.0118<br>Val = NA | Train = 96.50<br>Val = 99.12 | Train = 96.81<br>Val = 99.28 | Train = 96.52<br>Val = 99.13 |

**Table 4: Performance of different deep learning models on the dataset.**

*4.1.1. DenseNet:* DenseNet architectures showed strong performance across all metrics. DenseNet 169 achieved a training accuracy of 98.27% and a validation accuracy of 93.91%. Also, it had F1 scores of 95.22% in training and 93.95% in validation. Impressively, DenseNet 201 outperformed by a slightly better training accuracy of 98.80% and had F1-scores of 96.36% in training and 92.22% in validation, though its validation accuracy was somewhat lower at 92.17%.

Figure 6 shows training and validation set loss curves decreasing rapidly, finally reaching a plateau, indicating good learning and model convergence. The relatively smooth curve showed a minimal gap in Figure 7, indicating good generalization and no significant overfitting.

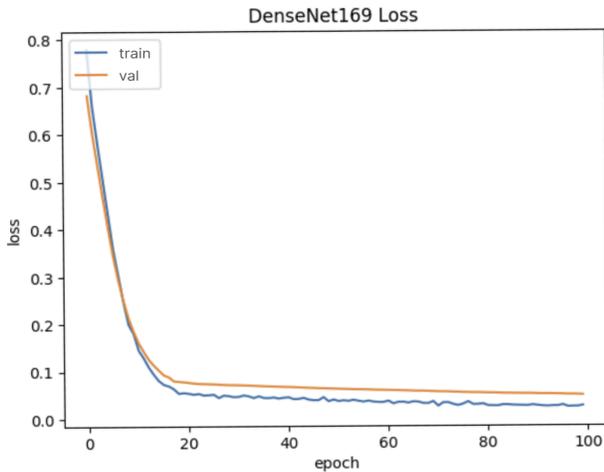

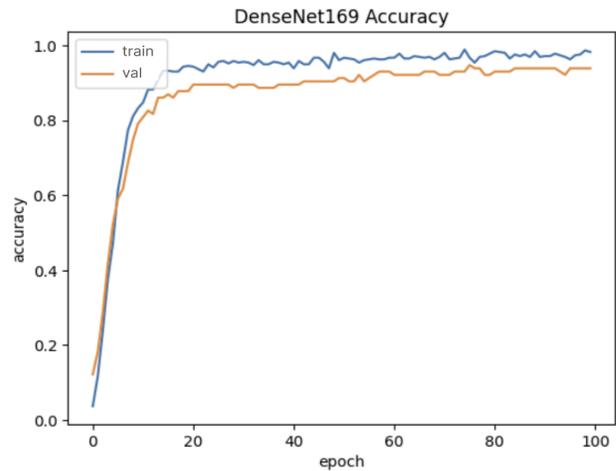

**Figure 6: Loss curve of DenseNet169.**

**Figure 7: Accuracy curve of DenseNet169.**

*4.1.2. ResNet:* The ResNets architecture also showed promising results. Resnet 101 V2 reached a training accuracy of 98.74% and a validation accuracy of 92.17%. Its F1 scores were 97.36% in training and 92.09% in validation, showing robust performance. Resnet 152 V2 showed a training accuracy of 98.20% and a validation accuracy of 93.04%, with high F1-scores of 95.79% in training and 93.22% in validation.

The training and validation data loss plunged drastically initially and finally tended towards low values, similar to DenseNet169, as shown in Figure 8. The very close train and validation loss curves indicated that the model generalized well without overfitting—a similar trend in the training and validation accuracy from Figure 9.

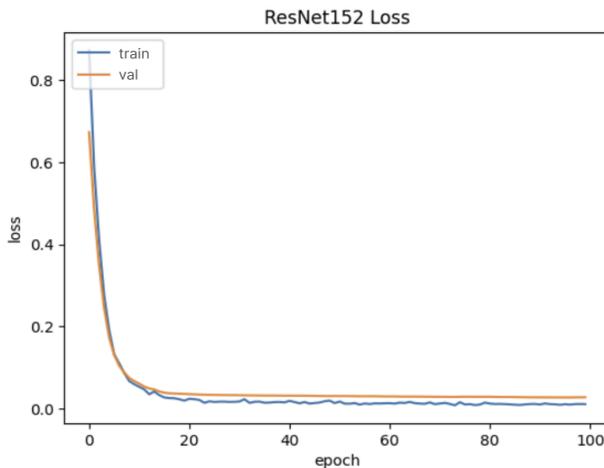

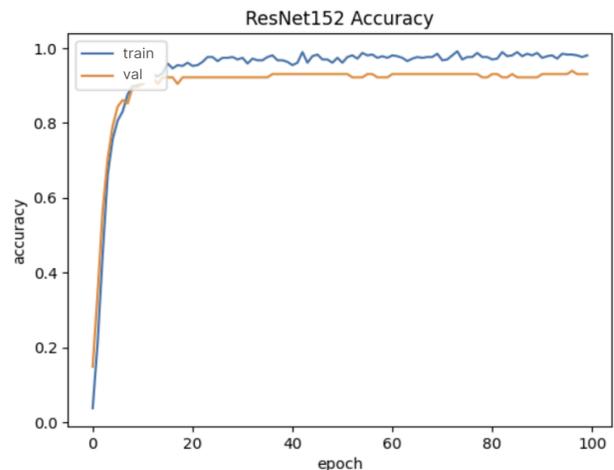

**Figure 8: Loss curve of ResNet152.**

**Figure 9: Accuracy curve of ResNet152.**

*4.1.3. VGG:* In contrast, the VGG architectures underperformed significantly compared to DenseNet and ResNet models. VGG 16 and VGG 19 showed much lower training accuracies (46.00% and 32.67%, respectively) and validation accuracies (33.91% and 33.04%, respectively). Their F1 scores were also considerably lower, with VGG 16 at 42.89% for training and 28.65% for validation and VGG 19 at 29.82% for training and 31.19% for validation.

The training loss decreased steadily, while the validation loss followed a different pattern, decreasing even slower and with apparent variance, as shown in Figure 10. There was a visible gap between training and validation loss, showing the possible overfitting phenomenon, i.e., a model fitting much better with the training set than the validation set. Also, Figure 11 shows oscillations in the training and validation accuracies.

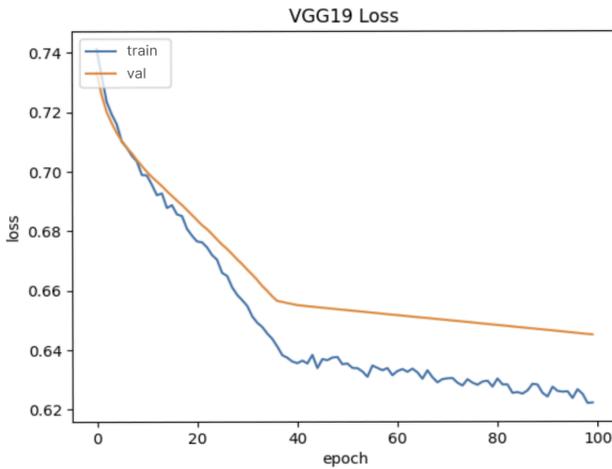
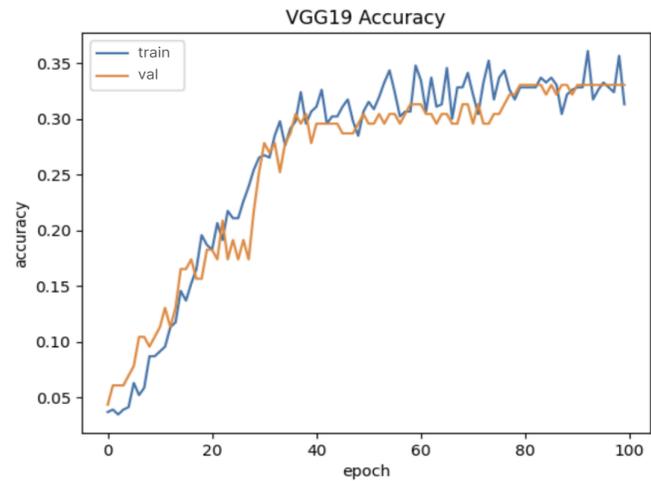

**Figure 10: Loss curve of VGG19.**  **Figure 11: Accuracy curve of VGG19.**

*4.1.4. YOLOV8:* The YOLOV8 model achieved an accuracy of 97.39% in training and 99.13% in validation with a shallow loss of 0.01175, which showed that despite being known for detection purposes, it surprisingly worked well for classification tasks. The F1 score was 96.5% in training and 99.12% in validation.

Figure 12 shows that the loss decreased rapidly and plateaued early in training, which shows the model's effectiveness for fast convergence during transfer learning. Unlike the loss, the accuracy oscillated slightly before it finally plateaued near 50 epochs, as indicated in Figure 13.

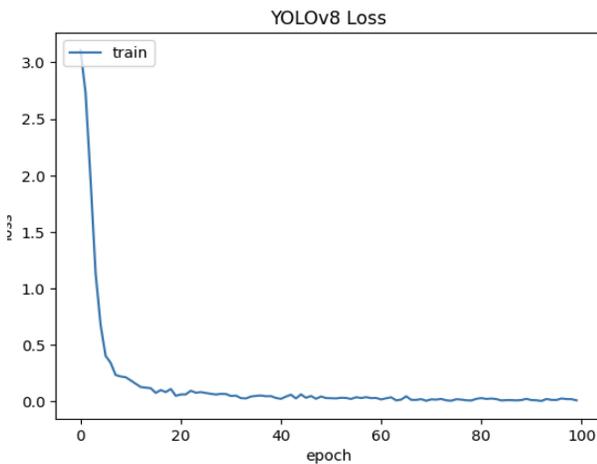
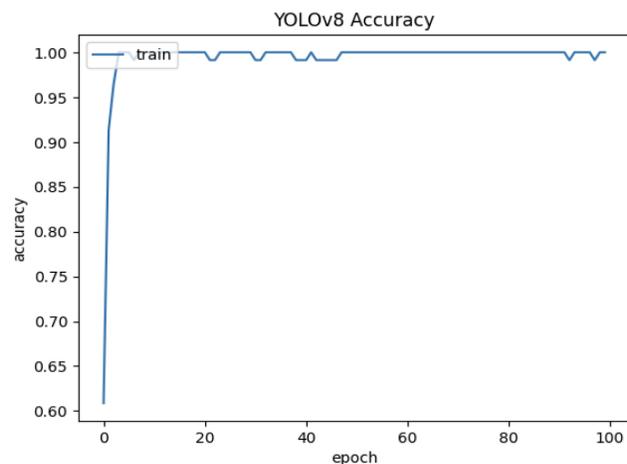

**Figure 12: Loss curve of YOLOv8.**  **Figure 13: Accuracy curve of YOLOv8.**

*4.1.5. CNN:* We validated different architectures for the vanilla CNN with the same dataset. However, we did not get satisfactory results. Since the dataset contained images of species across 23 classes, simple, shallow CNN could not converge effectively. Nevertheless, we tried adjusting various parameters like the number of layers, regularization techniques, loss functions, and augmentations. Despite these efforts, the performance metrics remained mediocre.

*4.2. Discussions*

The experimental analysis shows the metrics of various deep-learning models in identifying endangered animals on the custom endangered wildlife dataset, as shown in Figure 14. The dataset was carefully created from reputable online databases, ensuring the species' authenticity and relevance. The motto was to train a model that could recognize vulnerable species and assist in their proper monitoring. Furthermore, the other side of this study involved finding the best available architecture for this task. We experimented with various such architectures and analyzed their performance under different metrics.

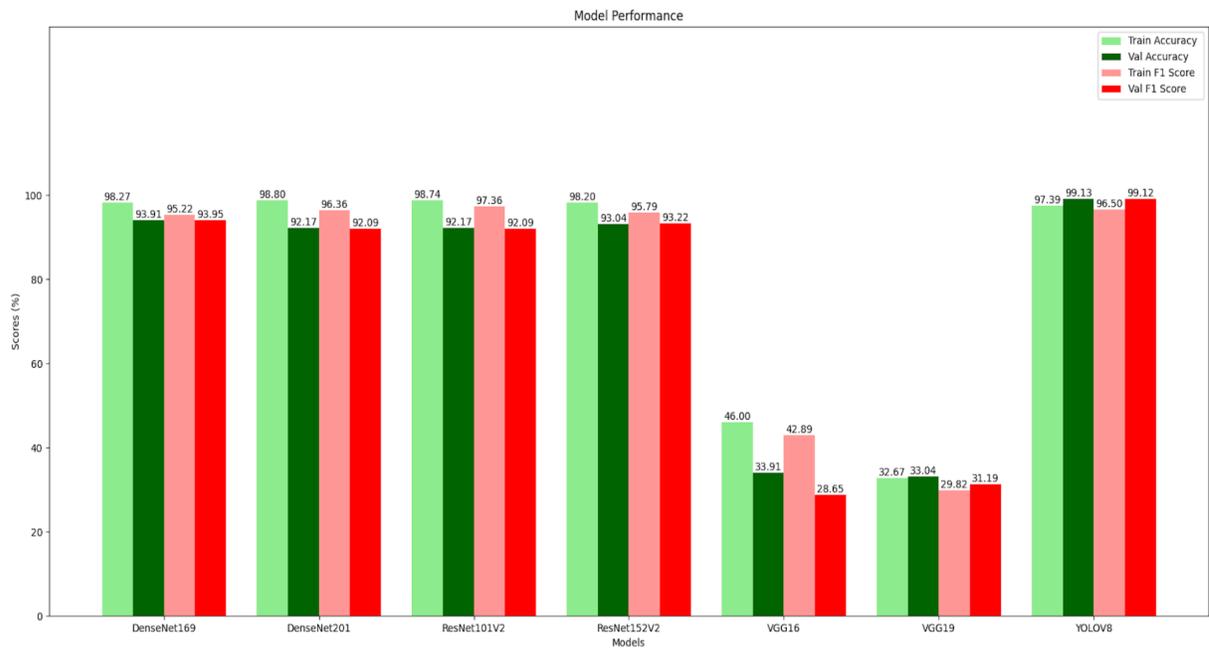

Figure 14: Comparison of metrics for different models.

We found that newer version of all the models showed stronger performance. VGG was the oldest among the models, so it performed poorly. Its accuracy was way below the vanilla CNN. Conversely, DenseNet and ResNet offered significantly better performance, so we can easily use them for related tasks. Also, we noted that these networks' newer and deeper versions did not provide any impactful difference across various metrics.

In addition, we implemented transfer learning and did not train them from scratch, benchmarking their ease of use in downstream tasks like this. We had to freeze the feature extraction layers and only trained the last few fully connected layers. Doing this saved the time and computing required without compromising their performance.

Alongside standard CNN models like ResNet, DenseNet, and VGGNet, specially designed for classification tasks, we also experimented with YOLOv8. YOLOs are primarily designed and used as go-to models for detection and segmentation tasks. To the great surprise, the metrics surpassed other standard classification models. Thus, YOLO might work well for classification tasks for custom datasets in similar niches.

YOLOv8's superior performance might be due to its advanced architecture, which combines efficient feature extraction with fast processing capabilities by integrating components like CSPNet (Cross Stage Partial Networks) and PANet (Path Aggregation Network) [29], [30]. CSPNet reduces the computational cost while maintaining accuracy. It divides the feature map into two parts and merges them through the cross-stage hierarchy. On the other hand, PANet enhances the information flow between various layers, which is excellent for object detection across multi-scales. But, this proved beneficial for classification tasks, too. Even more importantly, the multiscale capability of YOLOv8 allows handling images with object sizes that can have significant variations and differing resolutions. It also has advanced augmentations, such as mosaic augmentation, which places four training images into one with diverse contexts in one image and hence helps the model generalize better to varying lighting and environmental conditions in training. All these features, together, carry out fast processing and effective feature extraction.

## 5. Conclusion

To sum up, we performed experimental analysis on transfer learning of various deep-learning CNN architectures on the custom dataset containing images of endangered mammals from Nepal. The findings featured the superior performance of YOLOv8 compared to other models like DenseNet, ResNetV2, and VGG. It demonstrated higher accuracy, precision, and recall, making it practical for similar classification tasks. Although lagging by a narrow margin, other models like ResNet and DenseNet also performed well and competed neck and neck with YOLOv8. Transfer learning proved beneficial, drastically reducing training time and data required while maintaining high performance, which is crucial for tasks with limited data availability.

In the future, we will explore the ensemble methods that combine the strengths of multiple CNN architectures potentially enhancing classification accuracy and robustness, especially in diverse and challenging environmental conditions. We will also incorporate real-time monitoring capabilities in the future to provide feedback for conservation and take timely action in preventing the loss of endangered species. Hence, this study shows the reliance and robustness of deep-learning models in monitoring wildlife.

## 6. ACKNOWLEDGEMENT

The authors thank their supervisor, Dr. Mansi Bhavsar, for their invaluable guidance and support. Both authors contributed equally to this work.

# References

[1] S. Jnawali, H. Baral, S. Lee, et al., "The status of nepal mammals: The national red list series, department of national parks and wildlife conservation kathmandu, nepal," Preface by Simon M. Stuart Chair IUCN Species Survival Commission The Status of Nepal's Mammals: The National Red List Series, vol. 4, 2011.

[2] Y. LeCun, L. Bottou, Y. Bengio, and P. Haffner, "Gradient-based learning applied to document recognition," Proceedings of the IEEE, vol. 86, no. 11, pp. 2278–2324, 1998.

[3] O. Russakovsky, J. Deng, H. Su, et al., "Imagenet large scale visual recognition challenge," International journal of computer vision, vol. 115, pp. 211–252, 2015.

[4] C. Szegedy, W. Liu, Y. Jia, et al., "Going deeper with convolutions," in Proceedings of the IEEE conference on computer vision and pattern recognition, 2015, pp. 1–9.

[5] A. Krizhevsky, I. Sutskever, and G. E. Hinton, "Imagenet classification with deep convolutional neural networks," Advances in neural information processing systems, vol. 25, 2012.

[6] S. J. Pan and Q. Yang, "A survey on transfer learning," IEEE Transactions on knowledge and data engineering, vol. 22, no. 10, pp. 1345–1359, 2009.

[7] G. Huang, Z. Liu, L. Van Der Maaten, and K. Q. Weinberger, "Densely connected convolutional networks," in Proceedings of the IEEE conference on computer vision and pattern recognition, 2017, pp. 4700–4708.

[8] K. He, X. Zhang, S. Ren, and J. Sun, "Deep residual learning for image recognition," in Proceedings of the IEEE conference on computer vision and pattern recognition, 2016, pp. 770–778.

[9] K. Simonyan and A. Zisserman, "Very deep convolutional networks for large-scale image recognition," arXiv preprint arXiv:1409.1556, 2014.

[10] J. Redmon, S. Divvala, R. Girshick, and A. Farhadi, "You only look once: Unified, real-time object detection," in Proceedings of the IEEE conference on computer vision and pattern recognition, 2016, pp. 779–788.

[11] N. K. El Abbadi and E. M. T. A. Alsaadi, "An automated vertebrate animals classification using deep convolution neural networks," in 2020 International Conference on Computer Science and Software Engineering (CSASE), IEEE, 2020, pp. 72–77.

[12] A. G. Villa, A. Salazar, and F. Vargas, "Towards automatic wild animal monitoring: Identification of animal species in camera-trap images using very deep convolutional neural networks," Ecological informatics, vol. 41, pp. 24–32, 2017.

[13] M. Ibraheam, F. Gebali, K. F. Li, and L. Sielecki, "Animal species recognition using deep learning," in Advanced Information Networking and Applications: Proceedings of the 34th International Conference on Advanced Information Networking and Applications (AINA-2020), Springer, 2020, pp. 523–532.

[14] J. Li, J. Chen, B. Sheng, et al., "Automatic detection and classification system of domestic waste via multimodel cascaded convolutional neural network," IEEE Transactions on Industrial Informatics, vol. 18, no. 1, pp. 163–173, 2022. doi: 10.1109/TII.2021.3085669.

[15] C.-A. Brust, T. Burghardt, M. Groenenberg, et al., "Towards automated visual monitoring of individual gorillas in the wild," in Proceedings of the IEEE International Conference on Computer Vision Workshops, 2017, pp. 2820–2830.

[16] S. Beery, G. Van Horn, and P. Perona, "Recognition in terra incognita," in Proceedings of the European conference on computer vision (ECCV), 2018, pp. 456–473.

[17] A. Yılmaz, G. N. Uzun, M. Z. Gürbüz, and O. Kıvrak, "Detection and breed classification of cattle using yolo v4 algorithm," in 2021 International Conference on INnovations in Intelligent SysTems and Applications (INISTA), IEEE, 2021, pp. 1–4.

[18] M. Kumar, A. K. Yadav, M. Kumar, and D. Yadav, "Bird species classification from images using deep learning," in International Conference on Computer Vision and Image Processing, Springer, 2022, pp. 388–401.

[19] H. Nguyen, S. J. Maclagan, T. D. Nguyen, et al., "Animal recognition and identification with deep convolutional neural networks for automated wildlife monitoring," in 2017 IEEE international conference on data science and advanced Analytics (DSAA), IEEE, 2017, pp. 40–49.

[20] iNaturalist, iNaturalist, https://inaturalist.org/, 2022.

[21] ZooChat, ZooChat: The world's largest community of zoo and animal conservation enthusiasts, https://www.zoochat.com/community/, 2002.

[22] C. Caldas, split-folders (version 0.5.1), Available at: https://pypi.org/project/split-folders/, 2021.

[23] M. A. A. Babu, S. K. Pandey, D. Durisic, A. C. Koppisetty, and M. Staron, "Impact of image data splitting on the performance of automotive perception systems," in International Conference on Software Quality, Springer, 2024, pp. 91–111.

[24] C. Shorten and T. M. Khoshgoftaar, "A survey on image data augmentation for deep learning," Journal of big data, vol. 6, no. 1, pp. 1–48, 2019.

[25] J. Lu, L. Tan, and H. Jiang, "Review on convolutional neural network (cnn) applied to plant leaf disease classification," Agriculture, vol. 11, no. 8, p. 707, 2021.

[26] S. Sharma, R. Ghimire, S. Gurung, and S. GC, "Gantavya: A landmark recognition system," 2024.

[27] V. H. Phung and E. J. Rhee, "A deep learning approach for classification of cloud image patches on small datasets," Journal of information and communication convergence engineering, vol. 16, no. 3, pp. 173–178, 2018.

[28] D. Reis, J. Kupec, J. Hong, and A. Daoudi, "Real-time flying object detection with yolov8," arXiv preprint arXiv:2305.09972, 2023.

[29] C.-Y. Wang, H.-Y. M. Liao, Y.-H. Wu, P.-Y. Chen, J.-W. Hsieh, and I.-H. Yeh, "Cspnet: A new backbone that can enhance learning capability of cnn," in Proceedings of the IEEE/CVF conference on computer vision and pattern recognition workshops, 2020, pp. 390–391.

[30] S. Liu, L. Qi, H. Qin, J. Shi, and J. Jia, "Path aggregation network for instance segmentation," in Proceedings of the IEEE conference on computer vision and pattern recognition, 2018, pp. 8759–8768